\documentclass[11pt,a4paper]{article}
\usepackage{times}
\usepackage{latexsym}
\usepackage[T1]{fontenc}

\usepackage[hyperref,acceptedWithA]{tacl2018v2}

\usepackage{microtype}

\usepackage{xspace,mfirstuc,tabulary}

\newif\iftaclinstructions
\taclinstructionsfalse 
\iftaclinstructions
\renewcommand{\confidential}{}
\renewcommand{\anonsubtext}{(No author info supplied here, for consistency with
TACL-submission anonymization requirements)}
\newcommand{\instr}
\fi

\iftaclpubformat 

\else

\fi

\usepackage{xspace}
\usepackage{comment}
\usepackage{amsmath}
\usepackage{amssymb}
\usepackage[shortlabels]{enumitem}
\usepackage{adjustbox}
\usepackage{booktabs}
\usepackage{mathtools}
\usepackage{wasysym}

\usepackage{footmisc}

\DeclareSymbolFont{extraup}{U}{zavm}{m}{n}
\DeclareMathSymbol{\varheart}{\mathalpha}{extraup}{86}
\DeclareMathSymbol{\vardiamond}{\mathalpha}{extraup}{87}

\newcommand{\bamboo}{\textsc{Bamboo\includegraphics[trim={0 6cm 0 0}, clip, scale=0.005]{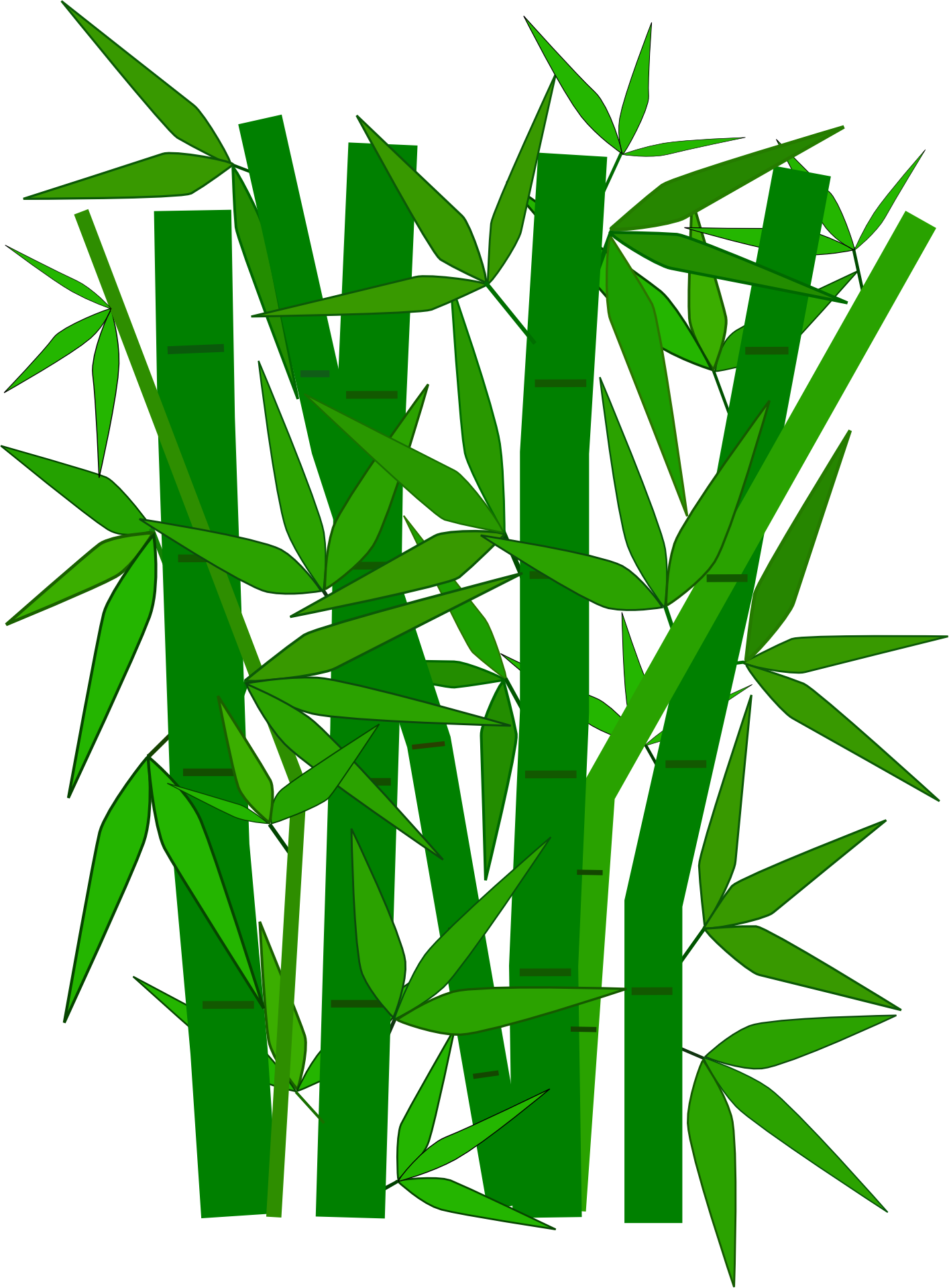}}\xspace}

\newcommand{\RC}{\textbf{Arg\lightning}\xspace}
\newcommand{\SYNO}{\textbf{Syno\lightning}\xspace}
\newcommand{\RFY}{\textbf{Reify\lightning}\xspace}

\newcommand{\SemBleu}{\textsc{SemBleu}\xspace}
\newcommand{\Sema}{\textsc{Sema}\xspace}
\newcommand{\Smatch}{\textsc{Smatch}\xspace}
\newcommand{\WSmatch}{\textsc{WSmatch}\xspace}
\newcommand{\SSmatch}{\textsc{S$^2$match}\xspace}

\usepackage{pifont}
\newcommand{\xmark}{\ding{55}}%

\title{Weisfeiler-Leman in the \textsc{Bamboo}\includegraphics[trim={0 6cm 0 0}, clip, scale=0.009]{figs/drawing-green-bamboo.png}: Novel AMR Graph Metrics \\ 
and a Benchmark for AMR Graph Similarity}

\author{Juri Opitz$^1$\quad Angel Daza$^{2}$\quad Anette Frank$^{1}$\quad \medskip\\
  $^1$Dept.\ of Computational Linguistics, Heidelberg University\\
  $^2$CLTL, Vrije Universiteit Amsterdam\\
  \medskip
  \texttt{\{opitz, frank\}@cl.uni-heidelberg.de}, ~~ \texttt{j.a.dazaarevalo@vu.nl}}

\date{}

\begin{document}
\maketitle
\begin{abstract}
Several metrics have been proposed for assessing the similarity of (abstract) meaning representations (AMRs), but little is known about how they relate to human similarity ratings.
Moreover, the current metrics have complementary strengths and weaknesses: some emphasize speed, while others make the alignment of graph structures explicit, at the price of a costly alignment step.  

In this work we propose new \textit{Weisfeiler-Leman AMR similarity metrics} that unify the strengths of previous metrics, while mitigating their weaknesses. Specifically, our new metrics are able to match contextualized substructures and induce n:m alignments between their nodes. Furthermore, we introduce a \underline{B}enchmark for \underline{A}MR \underline{M}etrics \underline{b}ased on \underline{O}vert \underline{O}bjectives (\bamboo), the first benchmark to support empirical assessment of graph-based MR similarity metrics. \bamboo maximizes the interpretability of results by defining multiple \textbf{overt objectives} that range from \textit{sentence similarity objectives} to \textit{stress tests} that probe a metric's robustness against meaning-altering and meaning-preserving graph transformations. We show the benefits of \bamboo by profiling previous metrics and our own metrics. Results indicate that our novel metrics may serve as a strong baseline for future work.

\end{abstract}

\section{Introduction}
Meaning representations aim at capturing the meaning of text in an explicit graph format. A prominent framework is abstract meaning representation \textit{(AMR}), proposed by \citet{banarescu-etal-2013-abstract}. AMR views sentences as rooted, directed, acyclic, labeled graphs. Their nodes are variables, attributes or (open-class) concepts and are connected with edges that express semantic relations.

There are many use cases in which we need to compare or relate two AMR graphs. A common situation is found in parser evaluation, where AMR metrics are widely applied \cite{may2016semeval, may2017semeval}.\footnote{With minor adaptions, AMR metrics are also used in other MR parsing tasks \cite{van-noord-etal-2018-evaluating, zhang-etal-2018-cross, oepen2020mrp}.}  Yet, there are more situations where we need to measure similarity of meaning as expressed in AMR graphs. E.g., \citet{bonial-etal-2020-infoforager} leverage AMR metrics in a semantic search engine for COVID-19  queries, \citet{naseem-etal-2019-rewarding} use metric feedback to reinforce AMR parsers, \citet{opitz-2020-amr} emulates metrics for referenceless AMR ranking and rating,  and \citet{opitz2020towards} employ AMR metrics for NLG evaluation. 

So far, multiple AMR metrics \cite{cai-knight-2013-smatch, cai-lam-2019-core, song-gildea-2019-sembleu, anchieta2019sema, opitz-tacl} have been proposed to assess AMR similarity. However, due to a lack of an appropriate evaluation benchmark, we have no empirical evidence that could tell us more about their strengths and weaknesses or offer insight about which metrics may be preferable over others in specific use cases.

Additionally, we would like to move beyond the aforementioned metrics and develop new metrics that account for graded similarity of graph substructures, which is
\begin{figure}
    \centering
    \includegraphics[width=0.75\linewidth]{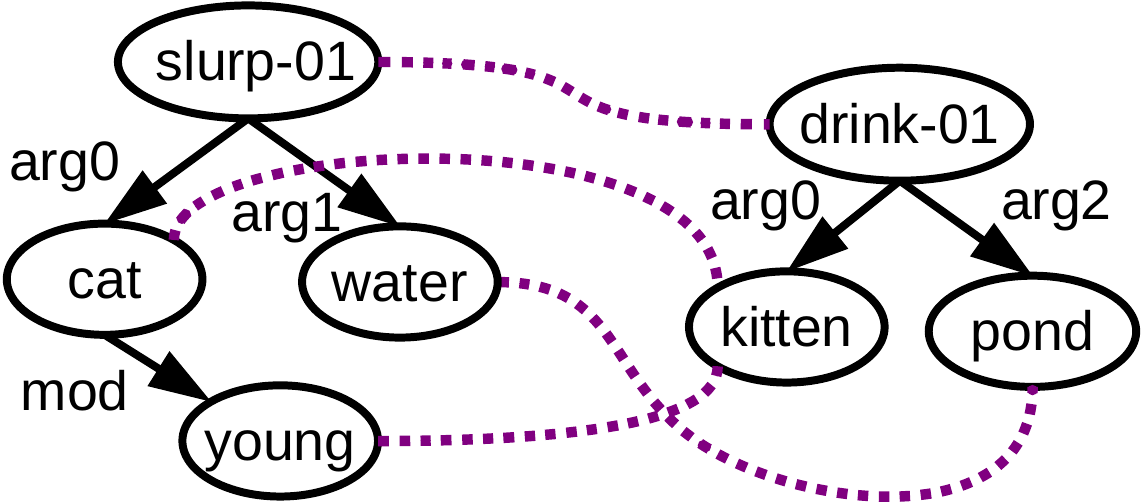}
    \caption{Similar AMRs, with sketched alignments.}
    \label{fig:ex1}
\end{figure}
 not an easy task. However, it is crucial  when we need to compare AMR graphs in a deeper way. Consider Fig.\ \ref{fig:ex1}, which shows two AMRs that convey very similar meanings. All aforementioned metrics assign this pair a low similarity score, and -- if alignment-based, as is \textsc{Smatch} \cite{cai-knight-2013-smatch} -- find only subpar alignments.\footnote{E.g., in Fig.\ \ref{fig:ex1}, \textsc{Smatch} aligns \textit{drink-01} to \textit{slurp-01} and \textit{kitten} to \textit{cat}, resulting in a single matching triple \textit{(x, arg0, y)}.} In this case, we want a metric that provides us with a high similarity score and, ideally, an explanatory alignment.

The structure of this paper is as follows. In \S \ref{sec:assess} we discuss related work. In \S \ref{sec:develop} we describe our first contribution: \textbf{new AMR metrics} that aim at unifying the strengths of previous metrics while mitigating their weaknesses. Specifically, our new metrics are capable of matching larger substructures and provide valuable n-m alignments in polynomial time. In \S \ref{sec:dat} we introduce \bamboo, our second contribution: it is \textbf{the first benchmark data set for AMR metrics} and includes novel robustness objectives that probe the behavior of AMR metrics under meaning-preserving and meaning-altering transformations of the inputs (\S \ref{sec:robustness}). In \S \ref{sec:exps}  we use \bamboo for a detailed, multi-faceted empirical study of previous and our proposed AMR metrics.

We release \bamboo and  our new metrics.\footnote{ \url{https://git.io/J0J7V}}
\label{sec:intro}

\section{Related work}
\label{sec:assess}

\paragraph{The classical AMR metric and its adaptions.} The `canonical' and widely applied AMR metric is \Smatch (\textit{S}emantic \textit{match}) \citep{cai-knight-2013-smatch}. It solves an NP-hard graph alignment problem approximately with a hill-climber and scores matching triples. \Smatch has been adapted  to \SSmatch (\textit{S}oft \textit{S}emantic \textit{match}), by \citet{opitz-tacl} to account for graded similarity of concept nodes (e.g., \textit{cat} -- \textit{kitten}), using word embeddings. \Smatch has also been adapted by \citet{cai-lam-2019-core} in \textit{W(eighted)Smatch} (\WSmatch), which penalizes errors relative to their distance to the root. This is motivated by the hypothesis that ``core semantics'' tend to be located near a graph's root. 

\paragraph{BFS-based and alignment-free AMR metrics} Recently, two new AMR metrics have been proposed: \Sema by \citet{anchieta2019sema} and \SemBleu by \citet{song-gildea-2019-sembleu}. Common to both is a mechanism that traverses the graph. Both start from the root, and collect structures with a breadth-first traversal (BFS). Also, both ablate the variable alignment of (W)\textsc{S$^{(2)}$match} and only consider their attached concepts, which increases computation speed. Apart from this, the metrics differ significantly: \SemBleu extracts bags of k-hop paths (k$\leq$3) from the AMR graphs and thereupon calculates BLEU \cite{papineni-etal-2002-bleu}. \Sema, on the other hand, is somewhat simpler and provides us with an F1 score that it achieves by comparing extracted triples. 
 
\paragraph{From measuring structure overlap to measuring meaning similarity} Most AMR metrics have been designed for semantic parser evaluation, and therefore determine a score for \textit{structure overlap}. While this is legitimate, with extended use cases for AMR metrics arising, there is increased awareness that structural matching of labeled nodes and edges of an AMR graph is not sufficient for assessing the \textit{meaning similarity} expressed by two AMRs \cite{kapanipathi2021leveraging}. This insufficiency has also been observed in cross-lingual AMR parsing
evaluation \cite{blloshmi-etal-2020-xl, sheth2021bootstrapping, uhrig-etal-2021-translate}, but is most prominent when attempting to compare the meaning of AMRs that represent different sentences \cite{opitz-tacl, opitz2020towards}. This work argues that in cases like Fig.\ \ref{fig:ex1}, the available metrics do not sufficiently reflect the similarity of the two AMRs and their underlying sentences.

\paragraph{How do humans rate similarity of sentence meaning?} STS \cite{baudivs2016sentence,baudis-etal-2016-joint,cer-etal-2017-semeval} and SICK \cite{marelli-etal-2014-sick} elicited human ratings of sentence similarity on a Likert scale. While STS annotates \textit{semantic} similarity, SICK annotates \textit{semantic} relatedness. These two aspects are highly related, but not the exact same \cite{budanitsky2006evaluating, kolb-2009-experiments}. Only the highest scores on the Likert scales of SICK and STS can be seen as reflecting the equivalence of meaning of two  sentences. Other data sets contain binary annotations of paraphrases \cite{dolan-brockett-2005-automatically}, that cover a wide spectrum of semantic phenomena.

\paragraph{Benchmarking Metrics} Metric benchmarking is an active topic in NLP research and led to the emergence of metric benchmarks in various areas, most prominently, MT and NLG \cite{gardent-etal-2017-webnlg, zhu2018texygen, wmt2019}. These benchmarks are useful since they help to assess and select metrics and encourage their further development \cite{gehrmann2021gem}. However, there is currently no established benchmark that defines a ground truth of \textit{graded semantic similarity between pairs of AMRs}, and how to measure it in terms of their structural representations. Also, we do not have an established ground truth to assess what alternative AMR metrics such as (W|S$^2$)\textsc{match} or \SemBleu really measure, and how their scores correlate with human judgments of the semantic similarity of sentences represented by AMRs. 

\section{Grounding novel AMR metrics in  Weisfleiler-Leman graph kernel }
\label{sec:develop}
Previous AMR metrics have complementary strengths and weaknesses. Therefore, we aim to propose new AMR metrics that are able to mitigate these weaknesses, while unifying their strengths, aiming at the best of all worlds. We want:
\vspace{-2mm}
\begin{enumerate}[label=\roman*),leftmargin=.3in, itemsep=-3.5ex]
\item  an \textbf{interpretable alignment} (\textit{\Smatch});\\
\item a \textbf{fast metric} (\textit{\Sema, \SemBleu});\\ 
\item \textbf{matching 
larger substructures} (\textit{\SemBleu}) \\
\item and \textbf{assessment of graded similarity of AMR subgraphs }(extending \textit{\SSmatch}).
\end{enumerate}
\vspace{-2mm}
This section proposes to make use of the  \textit{Weisfeiler-Leman graph kernel (WLK)} \cite{weisfeiler1968reduction, shervashidze2011weisfeiler} to assess AMR similarity. The idea is that WLK provides us with \SemBleu-like matches of larger sub-structures, while bypassing potential biases induced by the BFS-traversal \cite{opitz-tacl}. We then describe the \textit{Wasserstein Weisfeiler Leman kernel (WWLK)} \cite{NEURIPS2019_73fed7fd} that is similar to WLK but provides i) an alignment of atomic and non-atomic substructures (going beyond \Smatch) and ii) a graded match of substructures (going beyond \SSmatch). Finally, we further adapt WWLK to \textit{WWLK$_\Theta$}, a variant that we tailor to learn semantic edge parameters, to better assess AMR graphs.

\subsection{Basic Weisfeiler-Leman Kernel (WLK)}\label{subsec:wlk}
\begin{figure}
    \centering
    \includegraphics[width=\linewidth]{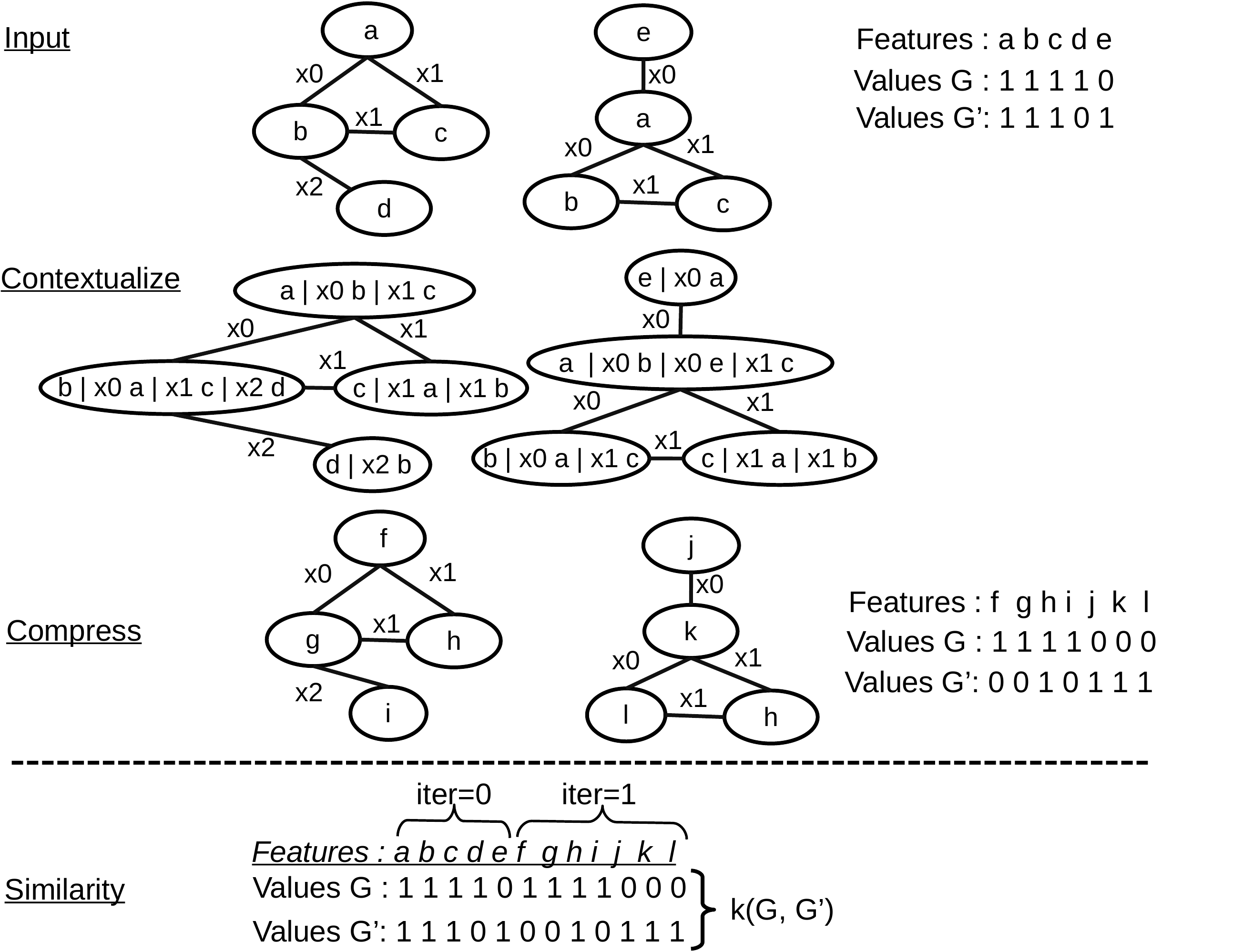}
    \caption{WLK example based on one iteration.}
    \label{fig:wlk-simple}
\end{figure}

The Weisfeiler-Leman kernel (WLK) method \cite{shervashidze2011weisfeiler} derives sub-graph features from two input graphs. WLK has shown its power in many tasks, ranging from protein classification to movie  recommendation \cite{NEURIPS2019_73fed7fd,yanardag2015deep}.  However, so far, it has not been applied to (A)MR graphs. In the following, we will describe the WLK method.

Generally, a kernel can be viewed as a similarity measurement between two objects \cite{hofmann2008kernel}, in our case, two AMR graphs $\mathcal{G}, \mathcal{G}'$. It is stated as $\mathbf{k}(\mathcal{G}, \mathcal{G}')=\langle\Phi(\mathcal{G}), \Phi(\mathcal{G}')\rangle$ where $\langle\cdot,\cdot\rangle: \mathbb{R}^d\times \mathbb{R}^d\rightarrow \mathbb{R}_+$ is an inner product and $\Phi$ maps an input to a feature vector that is built incrementally over $K$ iterations. For our AMR graphs, one such iteration $k$ works as follows: a) every node receives the labels of its neighbors and the labels of the edges connecting it to their neighbors, and stores them in a list (cf.\ \underline{Contextualize} in Fig.\ \ref{fig:wlk-simple}). b) The lists are alphabetically sorted and the string elements of the lists are concatenated to form new aggregate labels (cf.\ \underline{Compress} in Fig.\ \ref{fig:wlk-simple}). c) Two count vectors $x_{\mathcal{G}}^k$ and $x_{\mathcal{G}'}^k$ are created where each dimension corresponds to a node label that is found in any of the two graphs and contains its count (cf.\ \underline{Features} in Fig.\ \ref{fig:wlk-simple}). Since every iteration yields two vectors (one for each input), we can concatenate the vectors over iterations and calculate the kernel (cf.\ \underline{Similarity} in Fig.\ \ref{fig:wlk-simple}):

\begin{equation}
\begin{aligned}
\mathbf{k}(\cdot, \cdot) &= \langle\Phi_{WL}(\mathcal{G}),\Phi_{WL}(\mathcal{G}')\rangle\\ &= \langle concat(x_{\mathcal{G}}^0,...,x_\mathcal{G}^K), concat(x_{\mathcal{G}'}^0,...,x_\mathcal{G'}^K\rangle
\end{aligned}
\end{equation}

Specifically, we use the cosine similarity kernel and two iterations ($K$=2), which implies that every node receives information from its neighbors and their immediate neighbors. For simplicity we will first treat edges as undirected, but later will experiment with various directionality parameterizations.

\subsection{Wasserstein Weisfeiler-Leman (WWLK)}\label{subsec:wwlk}
\begin{figure}
    \centering
    \includegraphics[width=0.7\linewidth]{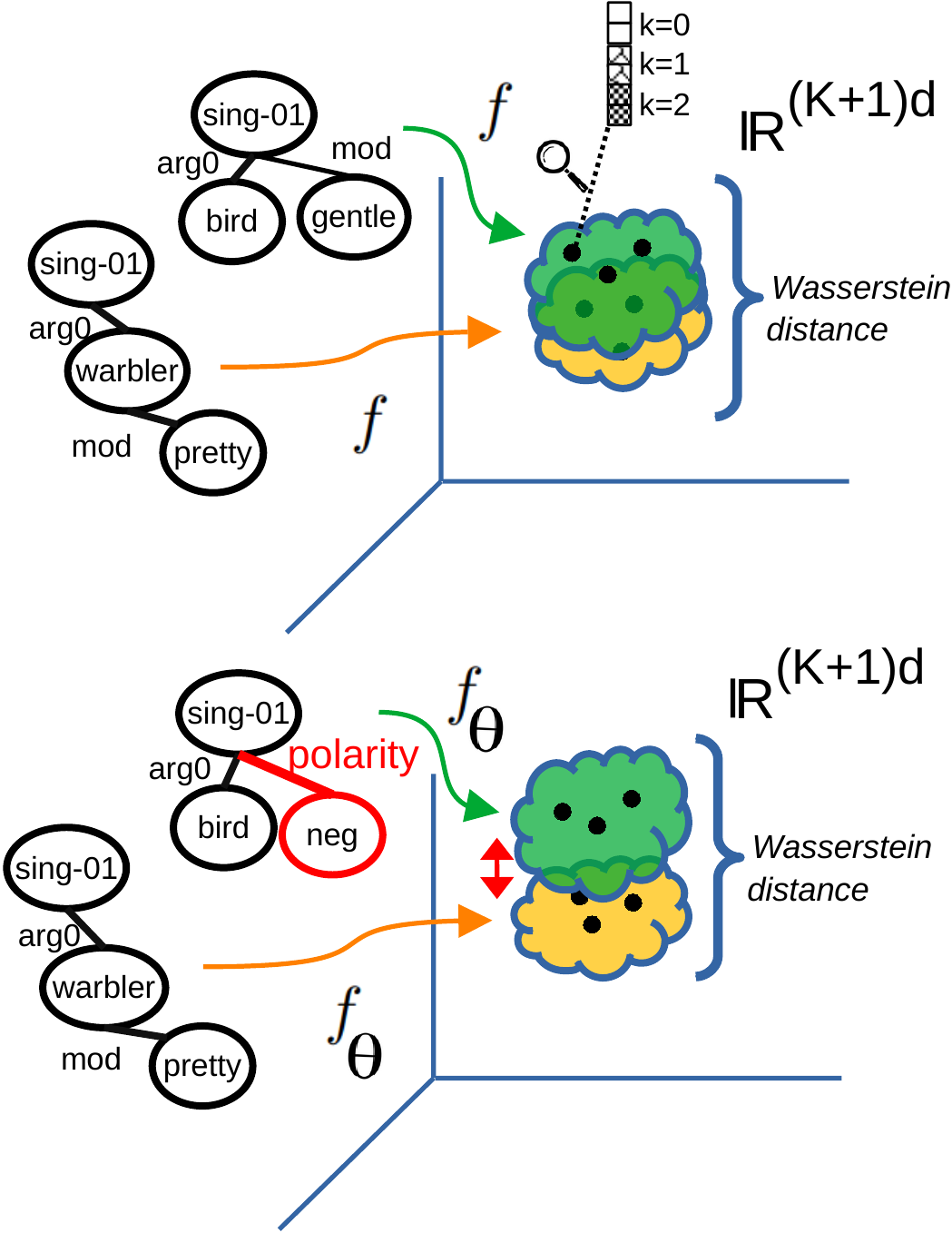}
    \caption{Wasserstein WLK example w/o learned edge parameters (top, \S \ref{subsec:wwlk}) and w/ learnt edge parameters (bottom, \S \ref{subsec:wwlkt}), which allow us to adjust the embedded graphs such that they better take the (impact of) AMR edges into account. Red: the distance increases because of a negation contrast between the two AMRs that otherwise convey similar meaning.}
    \label{fig:wlk-wasser}
\end{figure}

\SSmatch differs from all other AMR metrics in that it accepts close concept synonyms for alignment (up to a similarity threshold). But it comes with a restriction and a downside:
i) it cannot assess graded similarity of (non-atomic) AMR subgraphs, which is crucial for assessing partial meaning agreement between AMRs (as illustrated in Fig.\ \ref{fig:ex1}), and ii) the alignment is costly to compute. 

We hence propose to  adopt a variant of WLK: the Wasserstein-Weisfeiler Leman kernel (WWLK) \cite{NEURIPS2019_73fed7fd} for the following two reasons: i) WWLK can assess non-atomic subgraphs on a finer level, and ii) it provides graph alignments that are faster to compute than any of the existing \textsc{Smatch} metrics: \textsc{(W)S$^{(2)}$match}. 

WWLK works in \textbf{two steps}: 1. given its initial node embeddings, we use WL to project the graph into a \textit{latent space}, in which the final node embeddings describe \textit{varying degrees of contextualization}. 2. given a pair of such (WL) embedded graphs, a transportation plan is found that describes the minimum cost of transforming one graph into the other. In the top graph of Fig.\ \ref{fig:wlk-wasser}, $f$ indicates the first step, while \textit{Wasserstein distance} indicates the second. Now, we describe the steps in closer detail.

\paragraph{Step 1: WL graph projection into latent space} Let $v=1...n$ be the nodes of AMR $\mathcal{G}$. This graph is projected onto a matrix $\mathbb{R}^n \times \mathbb{R}^{(K+1)d}$ with

\begin{align}
 f(\mathcal{G}) &= hStack(X_\mathcal{G}^0,...,X_\mathcal{G}^K), \text{where} \\
    X_{\mathcal{G}}^k &= [x^k(1),...,x^{k}(n)]^T \in \mathbb{R}^n \times \mathbb{R}^d.
\end{align}

 $hstack$ concatenates matrices s.t.\ $(\big[\begin{smallmatrix}
  a & b\\
  c & d
\end{smallmatrix}\big], \big[\begin{smallmatrix}
  x & y\\
  w & z
\end{smallmatrix}\big]) \rightarrow \big[\begin{smallmatrix}
  a & b & x & y\\
  c & d & w & z
\end{smallmatrix}\big]$. This means that, in the output space, every node is associated with a vector that is itself a concatenation of $K+1$ vectors with $d$ dimensions each, where $k$ indicates the degree of contextualization (\includegraphics[scale=0.05]{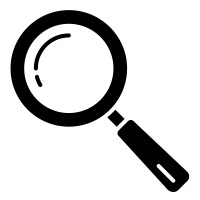} in Fig.\ \ref{fig:wlk-wasser}). The embedding  $x(v)^k \in \mathbb{R}^{d}$ for a node $v$ in a certain iteration $k$ is computed as follows:

\begin{align}
\label{eq:pass}
x(v)^{k+1} = \frac{1}{2} \bigg( x(v)^{k} + \frac{1}{d(v)} \sum_{u \in \mathcal{N}_v} w(u, v) \cdot x(u)^k\bigg).
\end{align}
$d(v)$ is the degree of a node, $\mathcal{N}$ returns the neighbors for a node, $w(u, v)$ can assign a weight to a node pair. The initial node embeddings, i.e., $x(\cdot)^0$, can be set up by looking up the node labels in a set of pre-trained word embeddings, or using random initialization. To distinguish between the discrete edge labels, we sample random weights. 

\paragraph{Step 2: Computing the Wasserstein distance between two WL-embedded graphs} The Wasserstein distance describes the minimum amount of work that is necessary to transform the (contextualized) nodes of one graph into the (contextualized) nodes of the other. It is computed based on pairwise euclidean distances from  $f(\mathcal{G})$ with $n$ nodes, and $f(\mathcal{G}')$ with $m$ nodes:
\begin{equation}\label{eq:emd}
distance = \sum_{i=1}^n \sum_{j=1}^m \mathbf{T}_{i,j}D_{i,j}
\end{equation}
Here, the `cost matrix' $D \in \mathbb{R}^{n \times m}$ contains the euclidean distances between the $n$ WL-embedded nodes from $\mathcal{G}$ and $m$ WL-embedded nodes from $\mathcal{G'}$. I.e., $D_{i,j} = ||f(\mathcal{G})_i - f(\mathcal{G}')_j||_2$. The \textit{flow matrix} $\mathbf{T}$ describes a transportation plan between the two graphs, i.e, $\mathbf{T}_{i,j}\geq 0$ states how much of node $i$ from $\mathcal{G}$ flows to node $j$ from $\mathcal{G}'$, the corresponding `local work' can be stated as $flow(i,j) \cdot cost(i,j) := \mathbf{T}_{i,j}\cdot D_{i,j}$. To find the best $\mathbf{T}$, i.e., the transportation plan that minimizes the cumulative work needed (Eq. \ref{eq:emd}), we solve a constraint linear problem:\footnote{We use \url{https://pypi.org/project/pyemd}}

\begin{align}\label{eq:optim}
    \min &\sum_{i=1}^n\sum_{j=1}^m \mathbf{T}_{i,j}D_{i,j}\\
    s.t.:~~& \mathbf{T}_{i,j}\ge0, 1\le i \le n, 1\le j \le m\\
    &\sum_{j=1}^m {\mathbf{T}_{i,j}} = \frac{1}{m}, 1 \le i \le n \\
    &\sum_{i=1}^n {\mathbf{T}_{i,j}} = \frac{1}{n}, 1 \le j \le m 
\end{align}
Note that i) the transportation plan $\mathbf{T}$ describes an n:m alignment between the nodes of the two graphs, and that ii) solving Eq.\ \ref{eq:optim} has polynomial time complexity, while the \textsc{(W)S$^{(2)}$match} problem is NP-complete.

\subsection{From WWLK to WWLK$_\theta$ with zero$^{th}$-order optimization}\label{subsec:wwlkt}

\paragraph{Motivation: AMR edge labels have meaning} The WL-embedding mechanism of WWLK (Eq.\ \ref{eq:pass}) associates a weight $w(u, v) \in \mathbb{R}$ with each edge. For unlabeled graphs, $w(u, v)$ is simply set to one. To distinguish between the discrete AMR edge labels, in WWLK we have used random weights. However, AMR edge labels encode complex relations between nodes, and simply choosing random weights may not be enough. In fact, we hypothesize that different edge labels may impact the meaning similarity of AMR graphs in different ways. Whereas a 
modifier relation in an AMR graph configuration
may or may not have a significant influence on the overall AMR graph similarity, an edge representing negation is bound to have a significant influence on the similarity of different AMR graphs. Consider the example in Fig.\ \ref{fig:wlk-wasser}: in the top figure, we  embed AMRs for \textit{The pretty warbler sings} and \textit{The bird sings gently}, which have similar meanings. In the bottom figure, the second AMR has been changed to express the meaning of \textit{The bird doesn't sing}, which clearly reduces the meaning similarity of the two AMRs. Hence, we hypothesize that learning edge parameters for different AMR relation types may help to better adjust the graph embeddings, such that the Wasserstein distance may increase or decrease, depending on the specific meaning of AMR relation labels, and thus to better capture global meaning differences between AMRs (as outlined in Fig.\ \ref{fig:wlk-wasser}: $f_\theta$).

Formally, to make the Wasserstein Weisfeiler-Leman kernel better account for \textit{edge-labeled} AMR graphs, we learn a parameter set $\Theta$ that consists of parameters $\theta^{edgeLabel}$, where $edgeLabel$ indicates the semantic relation, i.e., $edgeLabel \in L = \{\text{:arg0}, \text{:arg1},...,\text{:polarity},...\}$. 
Hence, in Eq.\ \ref{eq:pass}, we can set  $w(u, v)= \theta^{label(u, v)}$ and apply multiplication $\theta^{label(u, v)} \cdot x(u)^k$. To facilitate the multiplication, we  either may learn a matrix $\Theta \in \mathbb{R}^{|L| \times d}$ or a parameter vector $\Theta \in R^{|L|}$. In this paper, we constrain ourselves to the latter setting, i.e., our goal is to learn a parameter vector $\Theta \in R^{|L|}$.

\paragraph{Learning edge labels with direct feedback} To find suitable edge parameters $\Theta$, we propose a zero$^{th}$ order (gradient-free \cite{conn2009introduction}) optimization setup, which has the advantage that we can \textit{explicitly} teach our metric to better correlate with human ratings, optimizing the desired correlation objective without detours. In our case, we apply a simultaneous perturbation stochastic approximation (SPSA) procedure to estimate gradients \cite{spall1987stochastic, spall1998overview, wang2020overview}.\footnote{It improves upon a classic Kiefer-Wolfowitz approximation \cite{kiefer1952stochastic} by requiring, per gradient estimate, only $2$ objective function evaluations instead of $2n$.}

Let $sim(B, \Theta) = -WWLK_\Theta(B)$ be the similarity scores obtained from a (mini-)batch of graph pairs  ($B=[(\mathcal{G}_j, \mathcal{G}'_j),...]$) as provided by (parametrized) $WWLK$. Now, let $Y$ be the human reference scores. Then we design the loss function as $J(Y, \Theta) := 1 - correlation(sim(B, \Theta), Y)$. Further, let $\mu$ be coefficients that are sampled from a Bernoulli distribution. Then the gradient is estimated as follows:

\begin{equation}
    \label{eq:grad}
\widehat{\nabla}_\Theta = \frac{J(Y, \Theta + c\mu) - J(Y, \Theta - c\mu)}{2c\mu}.
\end{equation}
Finally, we can apply the common SGD learning rule: $\Theta^{t+1} = \Theta^{t} - \gamma\widehat{\nabla}_\Theta$. The learning rate $\gamma$ and $c$ decrease proportionally to $t$.

\section{\bamboo: Creating the first benchmark for AMR similarity metrics}
\label{sec:dat}
We now describe the creation of \bamboo, which aims to provide the first benchmark that allows researchers to empirically i) assess AMR metrics, ii) compare AMR metrics, and possibly iii) train AMR metrics.

\paragraph{Grounding AMR similarity metrics in human ratings of semantic sentence similarity} As main criterion for assessing AMR similarity metrics, we use human judgments of the meaning similarity of sentences underlying pairs of AMRs. A corresponding principle has been proposed by \citet{opitz-tacl}: a metric of pairs of AMR graphs $\mathcal{G}$ and $\mathcal{G}'$ that represent sentences $s$ and $s'$ should reflect human judgments of semantic sentence similarity and relatedness:
\vspace*{-5mm}

\begin{equation}
     amrMetric(\mathcal{G}, \mathcal{G}') \approx humanScore(s, s')
\end{equation}

\paragraph{Similarity Objectives}\label{par:targets} Accordingly, we select, as evaluation targets for AMR metrics, three notions of sentence similarity, which  have previously been operationalized in terms of human-rated evaluation datasets: i) the semantic textual similarity (STS) objective from  \citet{baudivs2016sentence,baudis-etal-2016-joint}; ii) the sentence relatedness objective (SICK) from \citet{marelli-etal-2014-sick}; iii) the paraphrase detection objective (PARA) by \citet{dolan-brockett-2005-automatically}.

Each of these three evaluation data sets can be seen as a set of pairs of sentences $(s_i,s'_i)$  with an associated score $humanScore(\cdot)$ that provides the human sentence relation assessment score reflecting \textit{semantic similarity} (STS), \textit{semantic relatedness} (SICK) and \textit{whether sentences are paraphrastic} (PARA). Hence, each of these data sets can be described as $\{(s_i, s'_i, humanScore(s_i, s'_i) = y_i)\}_{i=1}^n$. Both STS and SICK offer scores on Likert scales, ranging from \textit{equivalence} (max) to \textit{unrelated} (min), while PARA scores are binary, judging sentence pairs as being paraphrases (1), or not (0). We min-max normalize the Likert scale scores to the range $[0,1]$ to facilitate standardized evaluation. 

For \bamboo, we replace each pair $(s_i, s'_i)$ with their AMR parses: $(p_i$ = $parse(s_i)$, $p'_i$ = $parse(s'_i))$, transforming the data into $\{(p_i, p'_i, y_i)\}_{i=1}^n$. This provides the main partition of the benchmarking data for \bamboo, henceforth denoted as \textbf{Main}\footnote{The other partitions, which are largely based on this data, will be introduced in \S \ref{sec:robustness}.}. Statistics of \textbf{Main} are shown in Table \ref{tab:datastats}). The sentences in PARA are longer
compared to STS and SICK. The corresponding AMR graphs are, on average, much larger in number of nodes, but less complex with respect to the average  density.\footnote{The lower average density could be caused, e.g., by the fact that the PARA data is sampled from news sources, which means that the AMRs contain more named entity structures that usually have more terminal nodes.} 

\begin{table}
    \centering
    \scalebox{0.73}{
    \begin{tabular}{@{}lrrr|rrrr@{}}
    \toprule
    & & & &  \multicolumn{4}{c}{graph statistics} \\
         & data instances & \multicolumn{2}{c|}{(s.\ length)} & \multicolumn{2}{c}{\# nodes} & \multicolumn{2}{c}{density} \\
         \midrule
        source & train/dev/test & avg.\ & 50$^{th}$ & avg.\ & 50$^{th}$ & avg.\ & 50$^{th}$   \\
       STS  &5749/1500/1379 & 9.9 & 8  & 14.1 & 12 & 0.10 & 0.08 \\
       SICK & 4500/500/4927& 9.6 &  9 & 10.7 & 10 & 0.11 & 0.1 \\
       PARA & 3576/500/1275 & 18.9 & 19 & 30.6 & 30 & 0.04 &  0.04 \\
       \bottomrule
    \end{tabular}}
    \caption{\bamboo data set statistics of the \textbf{Main} partition. Sentence length (s.\ length, displayed for reference only) and graph statistics (average and median) are calculated on the training sets.}
    \label{tab:datastats}
\end{table}

\paragraph{AMR construction} We choose a strong parser that achieves high scores in the range of human-human inter-annotator agreement estimates in AMR banking: The parser yields 0.80-0.83 Smatch F1 on AMR2 and AMR3. The parser, henceforth denoted as T5S2S, is based on an AMR fine-tuned T5 language model \cite{T5} and produces AMRs in a sequence-to-sequence fashion.\footnote{\url{https://github.com/bjascob/amrlib}} It is on par with the current state-of-the-art that similarly relies on seq-to-seq \cite{xu-etal-2020-improving}, but the T5 backbone alleviates the need for massive MT pre-training. To obtain a better picture of the graph quality we perform manual quality inspections. 

\paragraph{Manual data quality assessment: three-way graph quality ratings}\label{par:dataassses} From each data set (SICK, STS, PARA) we randomly select 100 sentences and create their parses with T5S2S. Additionally, to establish a baseline, we also parse the same sentences with the GPLA parser of \citet{lyu-titov-2018-amr}, a neural graph prediction system that uses latent alignments (that reports 74.4 Smatch score on AMR2). This results in 300 GPLA parses and 300 T5S2S parses. A human annotator\footnote{The human annotator is a proficient English speaker and has worked several years with AMR.} inspects the (shuffled) sample and assigns three-way labels: \textit{flawed} -- an AMR contains critical errors that distort the meaning significantly; \textit{silver} -- an AMR contains small errors that can potentially be neglected; \textit{gold} -- an AMR is acceptable.

\begin{table}
    \centering
    \scalebox{0.83}{
    \begin{tabular}{@{}lllll@{}}
    \toprule
                 & Parser& \%gold$\uparrow$ & \%silver & \%flawed$\downarrow$  \\
                 \midrule
        STS &GPLA& 43[33,53] & 37[28,46] & 20[12,27] \\
         &\textbf{T5S2S}& 54[44,64]$\dagger\ddagger$ & 41[31,50] & 5[0,9]$\dagger\ddagger$ \\
         \midrule
        SICK &  GPLA& 38[28,47] & 49[39,59] & 13[6,19]   \\
         &\textbf{T5S2S}& 48[38,58]$\dagger$ & 47[37,57] & 5[0,9]$\dagger\ddagger$  \\
         \midrule
        PARA & GPLA &  9[3,14] & 52[43,62] & 39[29,48]    \\
         &\textbf{T5S2S}& 21[13,29]$\dagger\ddagger$ & 63[54, 73]$\dagger\ddagger$ & 16[8,23]$\dagger\ddagger$  \\
         \midrule
         \midrule
         ALL & GPLA &  30[25,35] & 46[40,52] & 24[19,29]    \\
         &\textbf{T5S2S}& 41[35,46]$\dagger\ddagger$  & 50[45,56]& 9[5,12]$\dagger\ddagger$  \\
        \bottomrule
    \end{tabular}}
    \caption{Three-way graph assessment. [x,y]: 95-confidence intervals estimated with bootstrap. $\dagger$ ($\ddagger$) significant improvement of T5S2S over GPLA with $p<0.05~(p<0.005)$.}
    \label{tab:dataquality}
\end{table}

Results in Table \ref{tab:dataquality} show that the quality of T5S2S parses is substantially better than the baseline in all three data sets. The percentage of excellent parses increases considerably (STS: +11pp, SICK: +10pp, PARA: +11pp) while the percentage of flawed parses drops notably (STS: -15pp, SICK: -8pp, PARA: -23pp). The increases in gold parses and decreases in flawed parses are significant in all data sets ($p<0.05$, 10,000 bootstrap samples of the sample means).\footnote{$\mathcal{H}_0$(gold): amount of gold graphs T5S2S $\leq$ amount of gold graphs GPLA; $\mathcal{H}_0$(silver): amount of silver graphs T5S2S $\leq$ amount of gold graphs GPLA; $\mathcal{H}_0$(flawed): amount of gold graphs T5S2S $\geq$ amount of gold graphs GPLA. }

\section{\bamboo: robustness challenges} 
\label{sec:robustness}

Besides benchmarking AMR metric scores against human ratings, we are also interested in assessing a metric's robustness under \textbf{meaning-preserving} and \textbf{-altering} graph transformations. Assume we are given any pair of AMRs from paraphrases. A small change in structure or node content can lead to two outcomes: the graphs still represent paraphrases, or they do not. We consider a metric 
to be robust if its ratings correctly reflect such changes. 

Specifically, we propose three transformation strategies. i) Reification (\RFY), which 
changes the graph's surface structure, but not its meaning; ii) Concept synonym replacement (\SYNO), which also preserves meaning and may or may not change the graph surface structure; iii) Role confusion (\RC), which applies small changes to the graph structure that do not preserve its meaning.

\subsection{Meaning-preserving transforms} 

Generally, given a meaning-preserving function $f$ of a graph, i.e.,
\begin{equation} 
\label{eq:mptransform}
    \mathcal{G} \equiv f(\mathcal{G}),
\end{equation}
it is natural to expect that a semantic similarity function over the pair of transformed AMRs nevertheless stays stable, and thus satisfies:

\begin{align} 
\label{eq:mp_equation}    
    metric(\mathcal{G}, \mathcal{G}') \approx metric(f(\mathcal{G}), f(\mathcal{G}')).
\end{align}

\begin{figure}
    \centering
    \includegraphics[width=0.75\linewidth]{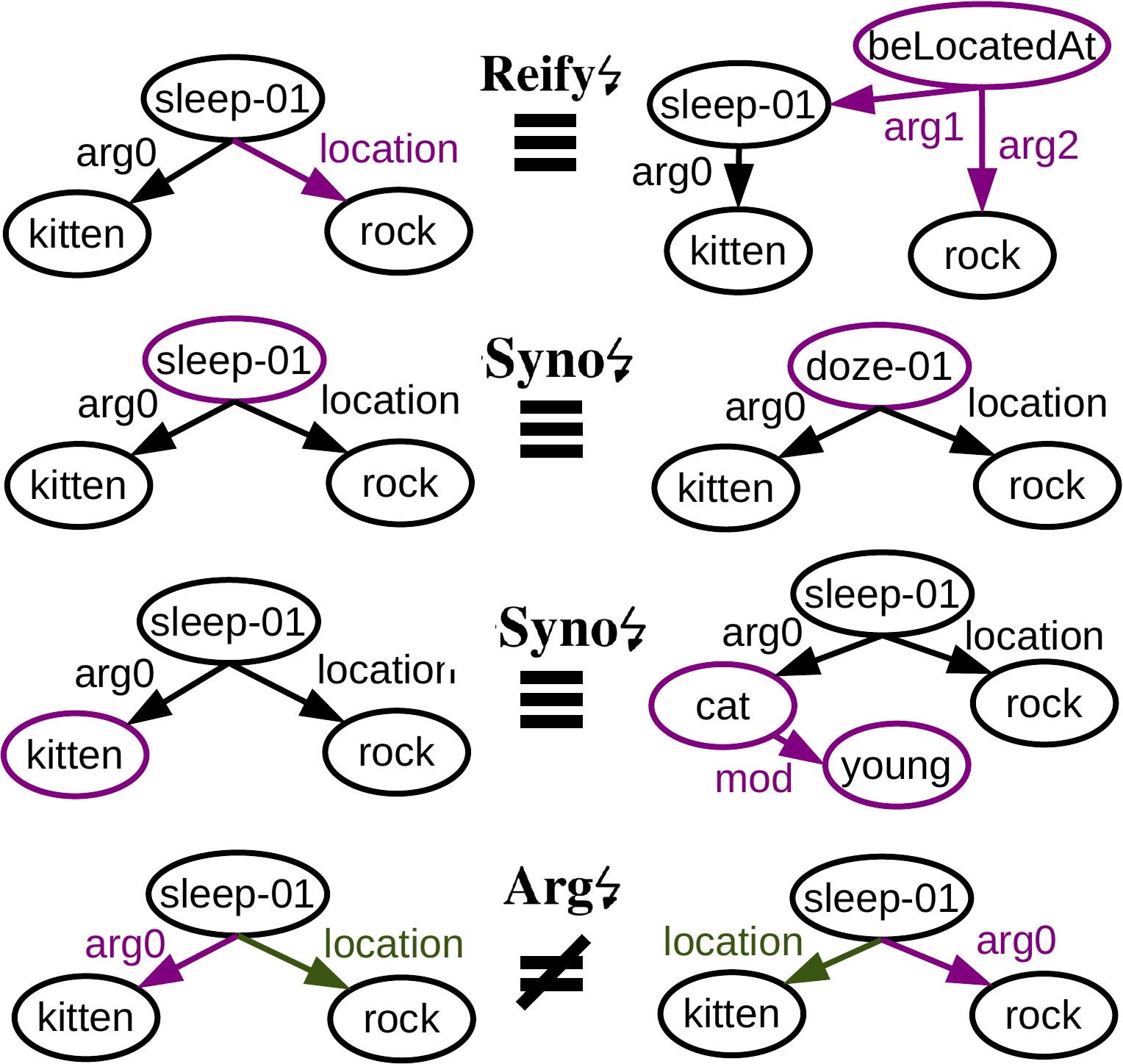}
    \caption{Examples for $f$ and $g$ graph transforms.}
    \label{fig:transforms}
\end{figure}

\paragraph{Reiification transform (\RFY)} \textit{Reification} is an established way to rephrase AMRs \citep{goodman-2020-penman}. Formally, a reification is induced by a rule
\vspace*{-6mm}

\begin{align}
edge(x, y) \xrightarrow{\text{reify}}{}&instance(z, h(edge)_0)\\ &\land h(edge)_1(z, x)\\ &\land h(edge)_2(z, y),
\end{align}
where $h$ returns, for a given edge, a new concept and corresponding edges from a dictionary, where the edges are either \texttt{:ARG}$_i$ or \texttt{:op}$_{i}$. An example is displayed in Fig.\ \ref{fig:transforms} (top, left). Besides reification for \textit{location}, other known types are \textit{polarity}-, \textit{modifier}-, or \textit{time}-reification.\footnote{A complete list of reifications are given in the official AMR guidelines: \url{https://github.com/amrisi/amr-guidelines/blob/master/amr.md}} Processing statistics of the applied reification operations are shown in Table \ref{tab:modification_ops}.

\begin{table}
    \centering
    \scalebox{0.73}{
    \begin{tabular}{@{}l|rr|rr|rr@{}}
    \toprule
         & \multicolumn{2}{c}{STS} & \multicolumn{2}{c}{SICK} & \multicolumn{2}{c}{PARA} \\
         \midrule
         & mean & th & mean & th & mean & th \\
        \RFY-OPS & 2.74 & [1, 2, 4] & 1.17 & [0, 1, 2] & 5.14 & [3, 5, 7]\\
        \SYNO-OPS & 0.80 & [0, 1, 2] & 1.31 & [0, 1, 2] & 1.30 & [0, 1, 2]\\
        \RC-OPS & 1.33 & [1, 1, 2] & 1.11 & [1, 1, 1]
 & 1.80 & [1, 2, 2]\\
 \bottomrule
    \end{tabular}}
    \caption{Statistics about the amount of transform operations that were conducted, on average, on one graph. [x,y,z]: 25th, 50th (median) and 75th percentile of the amount of operations.}
    \label{tab:modification_ops}
\end{table}

\paragraph{Synonym concept node transform (\SYNO)} Here, we iterate over AMR concept nodes. For any node that involves a predicate from PropBank, we consult a manually created database of (near-)synonyms that are also contained in PropBank, and sample one for replacement. E.g., some sense of \textit{fall} is near-equivalent to a sense of \textit{decrease} (\textit{car prices fell/decreased}). For concepts that are not predicates we run an ensemble of four WSD solvers\footnote{`Adapted lesk', `Simple Lesk', `Cosine Lesk', `max sim' \cite{banerjee2002adapted, lesk1986automatic, pedersen2007unsupervised}: \url{https://github.com/alvations/pywsd}.} (based on the concept and the sentence underlying the AMR) to identify its WordNet synset. From this synset we sample an alternative lemma.\footnote{To increase precision, we only perform this step if all solvers agree on the predicted synset.} If an alternative lemma consists of multiple tokens where modifiers precede the noun, we replace the node with a graph-substructure. So, if the concept is \textit{man} and we sample \textit{adult\_male}, we expand '$instance(x, man)$' with '$ mod(x, y) \land instance(y, adult) \land instance(x, male)$'. Data processing statistics are shown in Table \ref{tab:modification_ops}.

\begin{figure}
    \centering
    \includegraphics[width=0.9\linewidth]{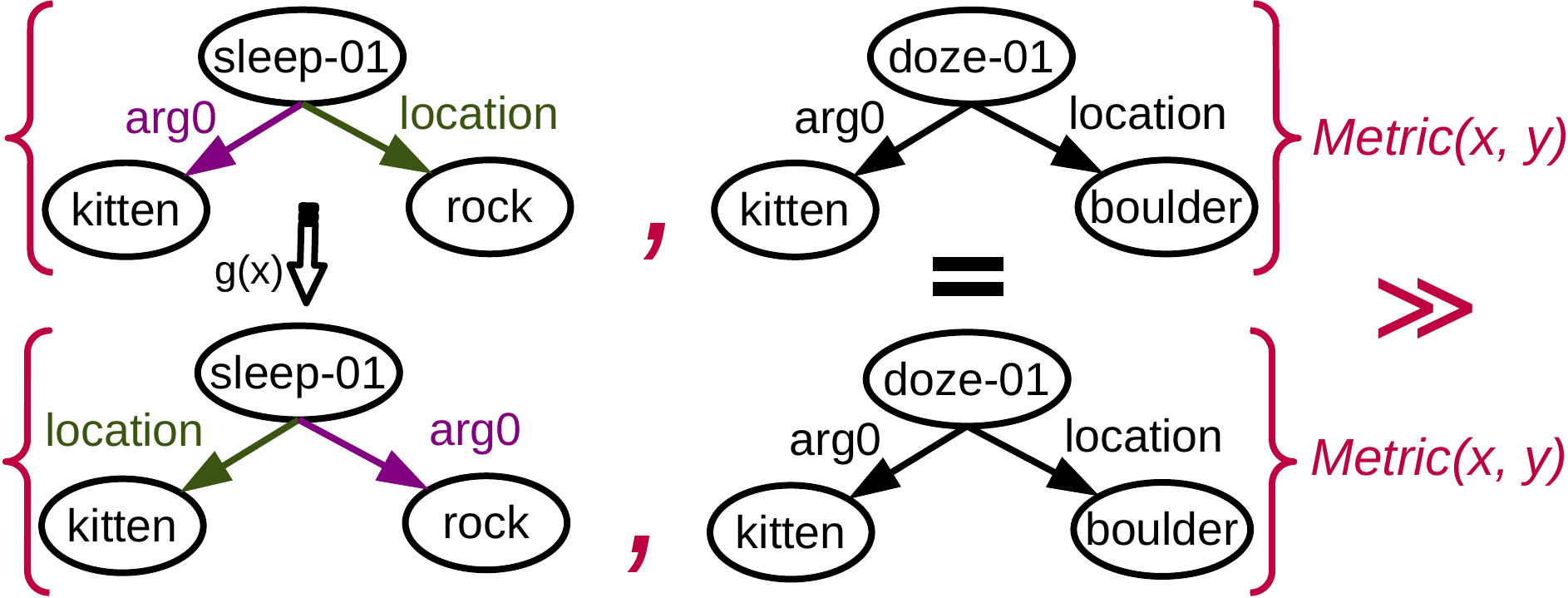}
    \caption{Metric objective example for \RC.}
    \label{fig:exampleg}
\end{figure}

\subsection{Meaning-altering graph transforms}

\paragraph{Role confusion (\RC)} A na\"ive AMR metric could be one that treats an AMR as a bag-of-nodes, omitting structural information, such as edges and edge-labels. Such metrics could exhibit misleadingly high correlation scores with human ratings, solely due to a high overlap in concept content.

Hence, we design adversarial instances that can probe an AMR metric when confronted with cases of opposing factuality (e.g., polarity, modality or relation inverses), while concept 
overlap is largely preserved. We design a function

\begin{equation}
\label{eq:matransform}
    \mathcal{G} \neq g(\mathcal{G}),
\end{equation}
that confuses role labels (see \RC in Figure \ref{fig:transforms}). We
make use of this function to turn two paraphrastic AMRs ($\mathcal{G}$, $\mathcal{G}'$) into non-paraphrastic AMRs, by appling $g$ to either $\mathcal{G}$, or $\mathcal{G}'$, but not both. 

In some cases $g$ may create a meaning that still makes sense (\textit{The tiger bites the snake.} $\rightarrow$ \textit{The snake bites the tiger.}), while in others, $g$ may induce 
a non-sensical meaning (\textit{The tiger jumps on the rock.} $\rightarrow$ \textit{The rock jumps on the tiger.}). However, this is not our primary concern, since in all cases, applying $g$ achieves our main goal: it returns a \textbf{different meaning} that turns a paraphrase-relation between two AMRs into a non-paraphrastic one.

To implement \RC, for each data set (\textbf{PARA}, \textbf{STS}, \textbf{SICK}) we create one new data subset. First, i) we collect all paraphrases from the initial data (in \textbf{SICK} and \textbf{STS} these are pairs with maximum human score).\footnote{This shrinks the train/dev/test size of STS (now: 474/106/158) and SICK (now: 246/50/238).} ii) We iterate over the AMR pairs $(\mathcal{G}, \mathcal{G}')$ and randomly select the first or second AMR from the tuple. We then collect all $n$ nodes with more than one outgoing edge. If $n=0$, we skip this AMR pair (the pair will not be contained in the data). If $n>0$, we apply the meaning altering function $g$ and randomly flip edge labels. Finally, we add the original $(\mathcal{G}, \mathcal{G}')$ to our data with the label \textit{paraphrase}, and the altered pair $(\mathcal{G}, g(\mathcal{G}'))$ with the label \textit{non-paraphrase} (cf.\ Figure \ref{fig:exampleg}). Per graph, we allow a maximum of 3 role confusion operations (see Table \ref{tab:modification_ops} for processing statistics).

\subsection{Discussion}

\paragraph{Safety of robustness objectives} We have proposed three challenging robustness objectives. \RFY changes the graph structure, but preserves the meaning. \RC keeps the graph structure (modulo edge labels) while changing the meaning. \SYNO changes node labels and possibly the graph structure and aims at preserving the meaning. 

\RFY and \RC are fully safe: they are well defined and are guaranteed to fulfill our goal (Eq.\ \ref{eq:mptransform} and \ref{eq:matransform}): meaning-preserving or -altering graph transforms. \SYNO is more experimental and has (at least) three failure modes. In the first mode, depending on context, a human similarity judgments could change when near-synonyms are chosen (sleep $\rightarrow$ doze, a young cat $\rightarrow$ kitten, etc.). The second mode occurs when WSD commits an error (e.g., minister (political sense) $\rightarrow$ priest). A third mode are societal biases found in WordNet (e.g., the node \textit{girl} may be mapped onto its `synonym' \textit{missy}). The third mode may not really be a failure, since it may not change the human rating, but, nevertheless, it may be undesirable.

In conclusion, \RFY and \RC confusion constitute safe robustness challenges, while results on \SYNO have to be taken with a grain of salt.

\paragraph{Status of the challenges in \bamboo and outlook} We believe that a key benefit of the robustness  challenges lies in their potential to provide complementary performance indicators, in addition to evaluation on the \textbf{Main} partition of \bamboo (cf.\ \S \ref{sec:dat}). In particular, the challenges may serve to assess metrics more deeply, uncover potential weak spots, and help select among metrics, e.g., when performance differences on \textbf{Main} are small. In this work, however, the complementary nature of \RFY, \SYNO or \RC versus \textbf{Main} is only reflected in the name of the partitions, and in our experiments, we consider all partitions equally. Future work may deviate from this setup.

Our proposed robustness challenges are also by no means exhaustive, and we believe that there is ample room for developing more challenges (\textit{extending} \bamboo) or experimenting with different setups of our challenges (\textit{varying} \bamboo\footnote{E.g., we may rectify only selected relations, or create more data, setting Eq.\ \ref{eq:mp_equation} to $metric(\mathcal{G}, \mathcal{G}') \approx metric(\mathcal{G}, f(\mathcal{G}'))$, only applying $f$ to one graph.}). For these reasons, it is possible that future work may justify alternative or enhanced setups, extensions and variations of \bamboo.

\section{Experimental insights}
\label{sec:exps}
\begin{table*}[t]
    \centering
    \scalebox{0.65}{
    \begin{tabular}{l|rr|rrr|rrr|rrr|rrr||rr}
     &  &  & \multicolumn{3}{c}{\textbf{Main}} & \multicolumn{3}{c}{\RFY} & \multicolumn{3}{c}{\SYNO} & \multicolumn{3}{c}{\RC} & amean & hmean\\
     & speed & align & STS & SICK & PARA & STS & SICK & PARA& STS & SICK & PARA& STS & SICK & PARA & - & -\\
     \midrule
     \Smatch &- & \checkmark &58.45 & 59.72 & 41.25 & 57.98 & 61.81 & 39.66 & 56.14 & 57.39 & 39.58 & 48.05 & 70.53 & 24.75 & 51.28 & 47.50 \\
     \midrule
     W\Smatch & -& \checkmark& 53.06 & 59.24 & 38.64 & 53.39 & 61.17 & 37.49 & 51.41 & 57.56 & 37.85 & 42.47 & 66.79 & 22.68 & 48.48 & 44.58 \\
    \SSmatch$_{default}$ & -& \checkmark& 56.38 & 58.15 & 42.16 & 55.65 & 60.04 & 40.41 & 56.05 & 57.17 & 40.92 & 46.51 & 70.90 & 26.58 & 50.91 & 47.80 \\
    \SSmatch & -& \checkmark& 58.82 & 60.42 & \textbf{42.55} & 58.08 & 62.25 & \textbf{40.60} & 56.70 & 57.92 & \textbf{41.22} & 48.79 & 71.41 & 27.83 & 52.22 & 49.07 \\
     \midrule
     \Sema & ++ & \xmark& 55.90 & 53.32 & 33.43 & 55.51 & 56.16 & 32.33 & 50.16 & 48.87 & 29.11 & 49.73 & 68.18 & 22.79 & 46.29 & 41.85 \\
       \SemBleu$_{k=1}$ &++ & \xmark& 66.03 & 62.88 & 39.72 & 61.76 & 62.10 & 38.17 & \textbf{61.83} & 58.83 & 37.10 & 1.99 & 1.47 & 1.40 & 41.11 & 5.78 \\
        \SemBleu$_{k=2}$ &++ & \xmark& 60.62 & 59.86 & 36.88 & 57.68 & 59.64 & 36.24 & 57.34 & 56.18 & 33.26 & 44.54 & 67.54 & 16.60 & 48.87 & 42.13\\
        \SemBleu$_{k=3}$ &++ & \xmark&  56.49 & 57.76 & 32.47 & 54.84 & 57.70 & 33.25 & 52.82 & 53.47 & 28.44 & 49.06 & 69.49 & 24.27 & 47.50 & 42.82\\
        \SemBleu$_{k=4}$ &++ & \xmark& 53.19 & 56.69 & 29.61 & 52.28 & 56.12 & 30.11 & 49.31 & 52.11 & 25.56 & 49.75 & 69.58 & 29.44 & 46.15 & 41.75 \\
        \midrule
        \midrule
        WLK (ours) & ++& \xmark & 64.86 & 61.52 & 37.35 & 62.69 & 62.55 & 36.49 & 59.41 & 56.60 & 33.71 & 45.89 & 64.70 & 19.47 & 50.44 & 44.35 \\
        WWLK (ours) & +& \checkmark& 63.15 & 65.58 & 37.55 & 59.78 & \textbf{65.53} & 35.81 & 59.40 & 59.98 & 32.86 & 13.98 & 42.79 & 7.16 & 45.30 & 28.83 \\
        WWLK$^{\Theta}$ (ours) & +& \checkmark& \textbf{66.94} & \textbf{67.64} & 37.91 & \textbf{64.34} & 65.49 & 39.23 & 60.11 & \textbf{62.29} & 35.15 & \textbf{55.03} & \textbf{75.06} & \textbf{29.64} & \textbf{54.90} & \textbf{50.26} \\
        \bottomrule
    \end{tabular}}
    \caption{\bamboo benchmark result of AMR metrics. All numbers are Pearson's $\rho \times 100$. ++: linear time complexity; +: polynomial time complexity; -: NP complete.}
    \label{tab:mainres}
\end{table*}

\paragraph{Questions posed to \bamboo}  \bamboo allows us to address several open questions: The first set of questions aims to gain more knowledge about previously released metrics. I.a., we would like to know: \textit{What semantic aspects of AMR does a metric measure? If a metric has hyper-parameters (e.g., \SemBleu), which hyper-parameters are suitable (for a specific objective)}? \textit{Does the costly alignment of \Smatch pay off, by yielding better predictions, or do the faster alignment-free metrics offer a `free-lunch'?} A second set of questions aims to evaluate our proposed novel AMR similarity metrics,  and to assess their potential advantages.

\paragraph{Experimental setup} We evaluate all metrics on the test set of \bamboo. The two hyper-parameters of \SSmatch, that determine when concepts are similar, are set with a small search on the development set (by contrast, \SSmatch$_{default}$ denotes the default setup). WWLK$_\theta$ is trained with batch size 16 on the training data. \SSmatch, WWLK and WWLK$_\theta$ all make use of GloVe embeddings \cite{pennington-etal-2014-glove}. 

Our main evaluation metric is Pearson's $\rho$ between a metric's output and the human ratings. Additionally, we consider two global performance measures to better rank AMR metrics: the arithmetic mean (\textit{amean}) and the harmonic mean (\textit{hmean}) over a metric's results achieved in all tasks. \textit{Hmean} is always $\leq$ $amean$ and is driven by low outliers. Hence, a large difference between amean and hmean serves as a warning light for a metric that is extremely vulnerable in a specific task.

\subsection{\bamboo studies previous metrics}
\label{subsec:oldmetrics}

Table \ref{tab:mainres} shows AMR metric results on \bamboo across all three human similarity rating types (STS, SICK, PARA) and our four challenges: \textbf{Main} repre\-sents the standard setup (cf.\ \S \ref{par:targets}), whereas \RFY, \SYNO and \RC test the metric robustness (cf.\ \S \ref{sec:robustness}).

\paragraph{\Smatch and \SSmatch rank 1$^{st}$ and 2$^{nd}$ 
of previous metrics} \textit{\Smatch}, our baseline metric, provides strong results across all tasks (Table \ref{tab:mainres}, amean: 51.28). With default parameters, \textit{\SSmatch}$_{default}$ performs slightly worse on the main data for STS and SICK, but improves upon \Smatch on PARA, achieving a slight overall improvement with respect to hmean (+0.30), but not amean (-0.37).  \SSmatch is more robust against  \SYNO (e.g., +4.6 on \SYNO STS vs.\ \Smatch), and when confronted with reified graphs (\RFY STS +3.3 vs.\ \Smatch). 

Finally, \textit{\SSmatch}, after setting its two hyper-parameters with a small search on the development data\footnote{STS/SICK: $\tau$=$0.90$, $\tau'$=$0.95$; PARA: $\tau$=$0.0, \tau'$=$0.95$}, consistently improves upon \Smatch over all tasks (amean: +0.94, hmean: +1.57).

\paragraph{W\Smatch: Are nodes near the root more important?} The hypothesis underlying W\Smatch is that concepts that are located near the top of an AMR have a higher impact on AMR similarity ratings. Interestingly, W\Smatch mostly falls short of \Smatch, offering substantially lower performance on all main tasks and all robustness checks, resulting in reduced overall amean and hmean scores (e.g., main STS: -5.39 vs.\ \Smatch, amean: -2.8 vs.\ \Smatch, hmean: -2.9 vs.\ \Smatch). This contradicts the `core-semantics' hypothesis and provides novel evidence that semantic concepts that influence human similarity ratings are not necessarily located close to AMR roots.\footnote{Manual inspection of examples shows that low similarity can frequently be explained with differences in concrete concepts that tend to be distant to the root. E.g., the low similarity (0.16) of \textit{\underline{Morsi} supporters clash with riot police in \underline{Cairo}}  vs.\  \textit{Protesters clash with riot police in \underline{Kiev}} arises mostly from \textit{Kiev} and \textit{Cairo} and \textit{Morsi}, however, these names (as are names in general in AMR) are distant to the root region, which is similar in both graphs (\textit{clash}, \textit{riot}, \textit{protesters}, \textit{supporters}).}

\paragraph{BFS-based metrics I: \textit{\Sema} increases speed but pays a price} Next, we find that \textit{\Sema} achieves lower scores in almost all categories, when compared with \Smatch (amean: -4.99, hmean -5.65), ending up at rank 7 (according to hmean and amean) among prior metrics. It is similar to \Smatch in that it extracts triples from graphs, but differs 
by not providing an alignment. Therefore, it can only loosely model some phenomena, and we conclude that the increase in speed comes at the cost of a substantial drop in modeling capacity.

\paragraph{BFS-based metrics II: \textit{\SemBleu} is fast, but is sensitive to $k$} 
Results for \textit{\SemBleu} show that it is very sensible to parameterizations of $k$. Notably, k=1, which means that the method only extracts bags of nodes, achieves strong results on SICK and STS. On PARA, however, \SemBleu is outperformed by \SSmatch, for all settings of $k$ (best $k$ (k=2): -2.8 amean, -4.7 hmean). Moreover, all variants of \SemBleu are vulnerable to 
robustness checks. E.g., k=2, and, naturally, k=1 are easily fooled by \RC, where  performance drops massively. k=4, on the other hand, is most robust against \RC, but overall it falls behind k=2. 

Since \SemBleu is asymmetric, we also re-compute the metric in a `symmetric' way by averaging the metric result over different argument orders. We find that this can slightly increase its performance (\textit{[k, amean, hmean]}: 
[1, +0.8, +0.6]; [2, +0.5, +0.4]; [3, +0.2, +0.2]; [4, +0.1, +0.0]).

In sum, our conclusions concerning \textit{\SemBleu} are: i) \SemBleu$_{k=1}$ (but not \SemBleu$_{k=3}$) performs well when measuring similarity and relatedness. However, \SemBleu$_{k=1}$ is na\"ive and easily fooled (\RC). ii) Hence, we recommend k=2 as a good tradeoff between robustness and performance, with overall rank 4 (amean) and 6 (hmean).\footnote{Setting k=2 stands in contrast to the original paper that recommended k=3, the common setting in MT. However, lower k in \SemBleu reduces biases \cite{opitz-tacl}, which may explain the better result on \bamboo.}

\subsection{\bamboo assesses novel metrics}
\label{subsec:newmetrics}

We now discuss results of our proposed metrics based on the Weisfeiler-Leman Kernel.

\paragraph{Standard Weisfeiler-Leman (WLK) is fast and a strong baseline for AMR similarity} First, we visit the classic Weisfeiler-Leman kernel. Like \SemBleu and \Sema, the (alignment-free) method is very fast. However, it outperforms these metrics in almost all tasks (score difference against second best alignment-free metric: ([a$\mid$h]mean: +1.6, +1.5) but falls behind  alignment-based \Smatch ([a$\mid$h]mean: -0.8, -3.2). Specifically, WLK proves robust against \RFY but appears more vulnerable against \SYNO (-5 points on STS and SICK) and \RC (notably PARA, with -10 points).\footnote{Similar to \SemBleu, we can mitigate this performance drop on \RC PARA by increasing the amount of passes $K$ in WLK, however, this decreases overall amean and hmean.}

The better performance, compared to \SemBleu and \Sema, may be due to the fact that WLK (unlike \SemBleu and \Sema) does not perform BFS traversal from the root, which may reduce biases.

\paragraph{WWLK and WWLK$_\theta$ obtain first ranks} 
Basic WWLK exhibits strong performance on SICK (ranking second on main and first on \RFY). 
However, it has large vulnerabilities, as exposed by \RC, where only \SemBleu$_{k=1}$ ranks lower. This can be explained by the fact that WWLK (7.2 Pearson's $\rho$ on PARA \RC) only weakly considers the semantic relations (whereas \SemBleu$_{k=1}$ does not consider semantic relations in the first place). 

WWLK$_\Theta$, our proposed algorithm for edge label learning, mitigates this vulnerability (29.6 Pearson's $\rho$ on PARA \RC, 1$^{st}$ rank). Learning edge labels also helps assessing similarity (STS) and relatedness (SICK), with substantial improvements over standard WWLK and \Smatch (STS: 66.94, +3.9 vs.\ WWLK and +10.6 vs.\ \Smatch; SICK +2.1 vs.\ WWLK and +8.4 vs.\ \Smatch). 

In sum, \textbf{WWLK$_\theta$ occupies rank 1 of all considered metrics} (amean and hmean), outperforming all non-alignment based metrics by large margins (amean +4.5 vs.\ WLK and +6.0 vs.\ \SemBleu$_{k=2}$; hmean +5.9 vs.\ WLK and +8.1 vs.\ \SemBleu$_{k=2}$), but also the alignment-based ones, albeit by lower margins (amean +2.7 vs.\ \SSmatch; hmean + 1.2 vs.\ \SSmatch).

\subsection{Analyzing hyper-parameters of (W)WLK}

\begin{table}
    \centering
    \scalebox{0.62}{
    \begin{tabular}{@{}l||r@{~~}r|r@{~~}r|r@{~~}r|r@{~~}r@{}}
    \toprule & \multicolumn{8}{c}{K (\#WL iters)}  \\
             & \multicolumn{2}{c}{basic (K=2)} & \multicolumn{2}{c}{K=1} & \multicolumn{2}{c}{K=3} & \multicolumn{2}{c}{K=4} \\
             & amean & hmean  & amean & hmean  & amean & hmean  & amean & hmean \\
         \midrule
         WLK & 50.4 & 44.4 & 49.8 & 44.2 & 47.6 & 42.4 & 46.4 & 41.5 \\
         WWLK & 45.3 & 28.8 &43.4 & 15.3 &  45.7 &31.4 & 42.3 & 24.0\\
         WWLK$_\theta$ & 54.9 & 50.3 & 52.2 & 35.4 & 55.2 & 51.1 & 50.8 & 47.3  \\
         \bottomrule
    \end{tabular}}
    \caption{WLK variants with different K.} 
    \label{tab:diffkweisf}
\end{table}

\paragraph{Setting K in (W)WLK} How does setting the number of iterations in Weisfeiler-Leman affect predictions? Table \ref{tab:diffkweisf} shows K=2 is a good choice for all WLK variants. K=3 slightly increases performance in the latent variants (WWLK: +0.4 amean; WWLK$_\theta$: +0.3 amean), but lowers performance for the fast symbolic matching WLK (-2.8 amean). This drop is somewhat expected: K$>$2 introduces much sparsity in the symbolic WLK feature space.

\begin{table}
    \centering
    \scalebox{0.64}{
    \begin{tabular}{@{}l||r@{~}r|r@{~~}r|r@{~~}r|r@{~~}r@{}}
    \toprule
    & \multicolumn{2}{c}{undirected} & \multicolumn{2}{c}{TOP-DOWN} & \multicolumn{2}{c}{BOTTOM-UP} & \multicolumn{2}{c}{2WAYS} \\
         &  amean &hmean &amean & hmean &amean & hmean&amean & hmean\\
         \midrule
        WLK & 50.4 & 44.4 & 50.3 & 44.3 & 50.2 & 43.8 & 49.5 & 41.8\\
        WWLK & 45.3 & 28.8 &  43.7 & 22.0 & 41.6 & 9.9 & 44.8 & 24.1\\
        WWLK$_\theta$ & 54.9 & 50.3 & 53.8 & 46.1 & 50.2 & 18.7 & 55.3 & 51.0 \\ 
        \bottomrule
    \end{tabular}}
    \caption{(W)WLK: 
    message passing directions.}
    \label{tab:direction}
\end{table}

\paragraph{WL message passing direction} Even though AMR defines directional edges, for optimal similarity ratings, it was not a-priori clear in which directions the node contextualization should be restricted when attempting to model human similarity.  Therefore, so far, our WLK variants have treated AMR graphs as undirected graphs ($\leftrightarrow$).  In this experiment, we study three 
alternate scenarios: `TOP-DOWN' (forward, $\rightarrow$), where information is only passed in the direction that AMR edges point at and `BOTTOM-UP' (backwards, $\leftarrow$), where information is exclusively passed in the opposite direction, and 2WAY ($\leftrightarrows$), where information is passed forwards, but for every edge $edge(x,y)$ we insert an $edge^{-1}(y, x)$. 2WAY facilitates more node interactions than either TOP-DOWN or BOTTOM-UP, while preserving directional information.

Our findings in Table \ref{tab:direction} show a clear trend: treating AMR graphs as graphs with undirected edges offers better results than TOP-DOWN (e.g., WWLK -1.6 amean; -6.6 hmean) and considerably better results when compared to WLK in BOTTOM-UP mode (e.g., WWLK -3.7 amean; -18.9 hmean). Overall, 2WAY behaves similarly to the standard setup, with a slight improvement for WWLK$_\theta$. Notably, the symbolic WLK variant, that does not use word embeddings, appears more robust in this experiment and differences between the three directional setups are small. 

\subsection{Revisiting the data quality in \bamboo} Initial quality analyses (\S \ref{par:dataassses}) suggested that the quality of \bamboo is high, with a large proportion of AMR graphs that are of gold or silver quality. In this experiment, we study how metric rankings and predictions could change when confronted with AMRs corrected by humans. From every data set, we randomly sample 50 AMR graph pairs (300 AMRs in total). In each AMR, the human annotator searched for mistakes, and corrected them.\footnote{Overall, few corrections were necessary, as reflected in a high \textsc{Smatch} between corrected and uncorrected graphs: 95.1 (STS), 96.8 (SICK), 97.9 (PARA).}

\begin{table}
    \centering
    \scalebox{0.65}{
    \begin{tabular}{@{}l|r@{~~}rr@{~~}rr@{~~}r|r@{~~}r@{}}
    \toprule
         & \multicolumn{2}{c}{STS} & \multicolumn{2}{c}{SICK} & \multicolumn{2}{c}{PARA} &  \multicolumn{2}{c}{AVERAGE} \\
         & MHA& IMA & MHA & IMA & MHA & IMA & MHA & IMA \\
         \midrule
        SM & [71, 73] & 97.9 & [66, 66] & 99.9 &  [44, 44] & 97.9 & [60, 61] & 98.6\\
        WSM & [64, 65]& 99.2 & [67, 67] & 99.8 & [47, 49] & 98.7 & [59, 60] & 99.2\\
        S2M$_{def}$ & [69, 70] & 97.7& [62, 63] & 99.3 & [44, 47] & 97.7 & [58, 60] & 98.2 \\
        S2M & [71, 73] & 97.8& [69, 70] & 98.6 & [41, 46] & 98.0 & [60, 63] & 98.1\\
        SE & [66, 66] & 97.7 & [55, 55] & 100 & [42, 46] & 99.0 & [55, 56] & 98.9 \\
        SB$_2$ & [68, 68] & 97.2 & [62, 62] & 99.8 & [41, 42]  & 98.8 & [57, 58] & 98.6 \\
        SB$_3$ & [66, 66] & 98.4 & [63, 63] & 99.7 & [33, 34]  & 99.3 & [54, 54] & 99.1 \\
        WLK & [72, 72] & 98.2 & [65, 65] & 99.8 & [43, 46] & 97.9 & [60, 61] & 98.6 \\
        WWLK & [77, 78] & 97.8 & [65, 67] & 98.1 & [42, 46] & 97.8 & [61, 63] & 97.9 \\
        WWLK$_\theta$ & [78, 78] & 96.8 & [67, 68] & 98.1 & [48, 48] & 96.7 & \textbf{[64, 65]} & 97.2\\
         \bottomrule
    \end{tabular}}
    \caption{Retrospective sub-sample quality analysis of \bamboo graph quality and sensitivity of metrics. All values are Pearson's $\rho \times 100$. Metric Human Agreement (MHA): $[x,y]$, where $x$ is the correlation (to human ratings) when the metric is executed on the uncorrected sample and $y$ is the same assessment on the manually post-processed sample. }
    \label{tab:dataquality2}
\end{table}

We study two settings. i) Intra metric agreement (IMA): For every metric, we calculate the correlation of its predictions for the initial graph pairs versus the predictions for the graph pairs that are ensured to be correct. Note that, on one hand, a \textit{high IMA for all metrics} would further corroborate the trustworthiness of \bamboo results. However, on the other hand, \textit{a high IMA for a single metric} cannot be interpreted as a marker for this metric's quality. I.e., a maximum IMA (1.0) could also indicate that a metric is completely insensitive to the human corrections. Furthermore, we study ii) Metric human agreement (MHA): Here, we correlate the metric scores against human ratings, once when fed the fully gold-ensured graph pairs and once when fed the standard graph pairs. Both measures, IMA, and IAA, can provide us with an indicator of how much metric ratings would change if \bamboo would be fully human corrected.

Results are shown in Table \ref{tab:dataquality2}. All metrics exhi\-bit high IMA, suggesting that potential changes in their ratings, when fed gold-ensured graphs, are quite small. Furthermore, on average, all metrics tend to exhibit slightly better correlation with the human when computed on the gold-ensured graph pairs. However, supporting the assessment of IMA, the increments in MHA appear small, ranging from a minimum increment of +0.3 (\SemBleu) to a maximum increment of +2.8 (\SSmatch), whereas WWLK yields an increment of +1.8. Generally, while this assessment has to be taken with a grain of salt due to the small sample size, it overall supports the validity of \bamboo results.

\subsection{Discussion}

\paragraph{Align or not align?} We can group metrics for  graph-based meaning representations into whether they compute an \textbf{alignment} between AMRs or not \cite{liu-etal-2020-dscorer}. A computed alignment, as in \Smatch, has the advantage that it lets us assess finer-grained AMR graph similarities and divergences, by creating and exploiting a mapping that shows which specific substructures of two graphs are more or less similar to each other. On the other hand, it was still an open question whether such an alignment is worth its computational cost and enhances similarity judgments. 

Experiments on \bamboo provide novel evidence on this matter: \textbf{alignment-based metrics may be preferred for better accuracy. Non-alignment based metrics may be preferred if speed matters most.} The latter situation may occur, e.g., when AMR metrics must be executed over a large cross-product of parses (for instance, to semantically cluster sentences from a corpus). For a balanced approach, WWLK$_\Theta$ offers a good trade-off: polynomial-time alignment and high accuracy. 

\begin{figure}
    \centering
   \includegraphics[width=0.76\linewidth]{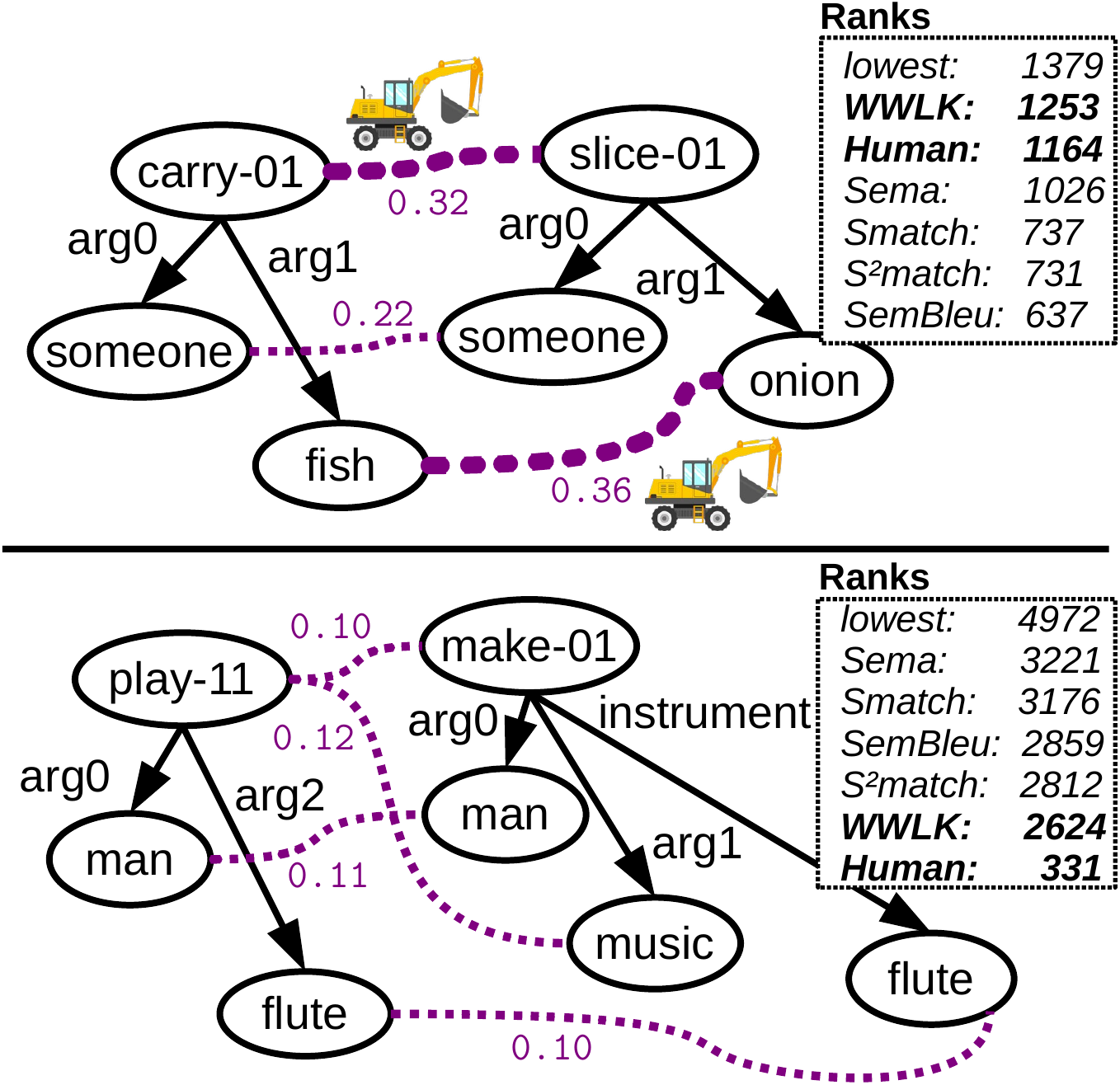}
    \caption{WWLK alignments 
    and metric scores for dissimilar (top, STS) and similar (bottom, SICK) AMRs. Excavators indicate heavy Wasserstein work $flow \cdot cost$.}
    \label{fig:sts-example}
\end{figure}

\paragraph{Example discussion I: Wasserstein transportation analysis explains disagreement} Fig.\ \ref{fig:sts-example} (top) shows an example where the human-assigned similarity score is relatively low (rank 1164 of 1379). Due to the graphs having the same structure ($x~arg0~y;~x~ arg1~z$), the previous metrics (except \Sema) tend to assign similarities that are relatively too high. In particular, \SSmatch finds the exact same alignments in this case, but cannot assess the concept-relations more deeply. WWLK yields more informative alignments since they explain its decision to assign a more appropriate lower rank (1253 of 1379): substantial work is needed to transport, e.g., \textit{carry-01} to \textit{slice-01}.

\paragraph{Example discussion II: the value of
n:m alignments} 

Fig.\ \ref{fig:sts-example} (bottom) shows that WWLK produces valuable n:m alignments (\textit{play-11} vs.\ \textit{make-01} and \textit{music}), which are needed to properly reflect similarity (note that \Smatch, \WSmatch and \SSmatch only provide 1-1 alignments). Yet, the example also shows that there is still a way to go. While humans assess this near-equivalence easily, providing a relatively high score (rank 331  of 4972), all metrics considered in this paper, including ours, assign relative ranks that are too low (WWLK: 2624). Future work may incorporate external PropBank \cite{palmer-etal-2005-proposition} knowledge into AMR metrics. In PropBank, sense \textit{11} of \textit{play} is defined as equivalent to \textit{making music}.

\section{Conclusion}

Our contributions in this work are three-fold: i) We propose a suite of novel Weisfeiler-Leman AMR similarity metrics that are able to reconcile a performance conflict between precision of AMR similarity ratings and the efficiency of computing alignments. ii) We release \bamboo, the first benchmark that allows researchers to assess AMR metrics empirically, setting the stage for future work on graph-based meaning representation metrics. iii) We showcase the utility of \bamboo,  by applying it to profile existing AMR metrics, uncovering hitherto unknown strengths or weaknesses, and to assess the strengths of our newly proposed metrics that we derive and further develop from the classic Weisfeiler-Leman Kernel. We show that through \bamboo we are able to gain novel insight regarding suitable hyperparameters of different metric types, and to gain novel perspectives on how to further improve AMR similarity metrics to achieve better correlation with the degree of meaning similarity of paired sentences, as perceived by humans.

\section*{Acknowledgements} We are grateful to three anonymous reviewers and Action Editor Yue Zhang for their valuable comments that have helped to improve this paper. We are also thankful to Philipp Wiesenbach for giving helpful feedback on a draft of this paper. This work has been partially funded by the DFG through the project ACCEPT as part of the Priority Program ``Robust Argumentation Machines'' (SPP1999).

\bibliography{acl2020}
\bibliographystyle{acl_natbib}

\end{document}